%% file: main.tex
\documentclass[10pt,twocolumn,letterpaper]{article}

\usepackage{iccv}              % To produce the CAMERA-READY version

\definecolor{iccvblue}{rgb}{0.21,0.49,0.74}
\usepackage[pagebackref,breaklinks,colorlinks,allcolors=iccvblue]{hyperref}

\usepackage{booktabs, xcolor, colortbl}
\usepackage{graphicx}
\usepackage{array}
\usepackage{amsmath}
\usepackage{siunitx}
\definecolor{darkgreen}{RGB}{0,175,0}
\newcommand{\increase}[1]{\textcolor{darkgreen}{\textuparrow{}}}
\newcommand{\decrease}[1]{\textcolor{red}{\textdownarrow{}}}
\usepackage{multirow}
\usepackage{algorithm,algorithmicx,algpseudocode}
\def\paperID{1801} % *** Enter the Paper ID here
\def\confName{ICCV}
\def\confYear{2025}
\usepackage[accsupp]{axessibility}  % Improves PDF readability for those with disabilities.
\title{ProbRes: Probabilistic Jump Diffusion for Open-World Egocentric Activity Recognition}

\author{Sanjoy Kundu$^*$, Shanmukha Vellamcheti\thanks{Equal Contribution}, Sathyanarayanan N. Aakur\\
CSSE Department, Auburn University\\
Auburn, Alabama, USA 36849\\
{\tt\small \{szk0266, szv0080, san0028\}@auburn.edu}
}

\begin{document}
\maketitle
\input{sec/0_abstract}    
\section{Introduction}

\begin{figure}[t]
    \centering
    \includegraphics[width=0.99\columnwidth]{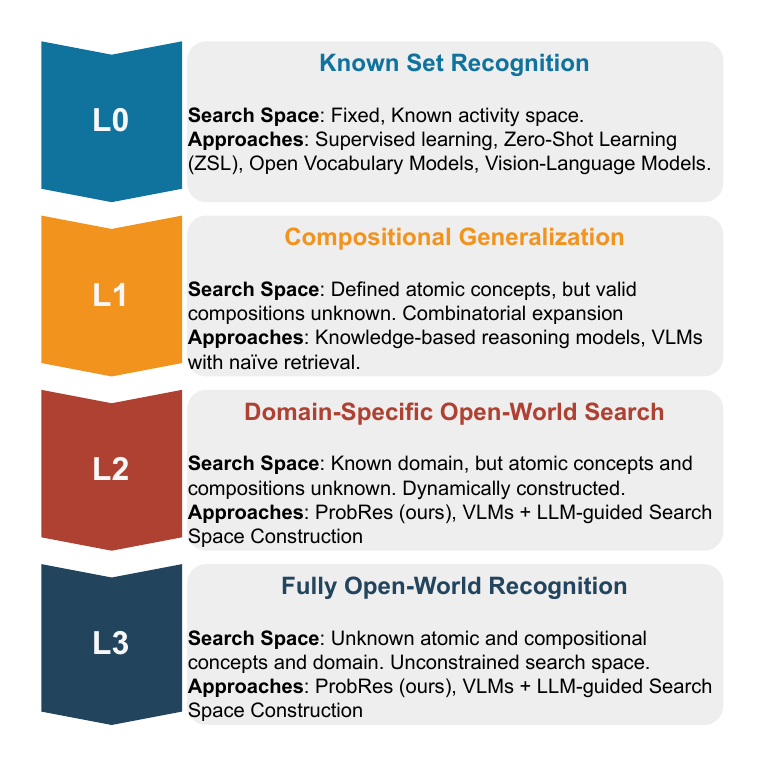}
    \caption{\textit{Taxonomy of Openness in Egocentric Activity Recognition.} We define four levels of openness based on search space constraints: L0 (fixed activity set), L1 (known atomic concepts, unknown compositions), L2 (known domain, inferred activities), and L3 (fully unconstrained search).
    % We define four levels of openness based on search space constraints. \textbf{L0}: Known-set recognition with predefined activities. \textbf{L1}: Compositional generalization, where atomic concepts are known but not compositions. \textbf{L2}: Domain-specific open-world search, where the domain is known, but everything else is inferred. \textbf{L3}: Fully open-world recognition, where neither the domain nor valid compositions are known. 
    % Our model \textit{ProbRes} enables efficient inference at L1-L3 through structured priors and probabilistic search.
    }
    \label{fig:taxonomy}
\end{figure}
Egocentric activity recognition in open-world settings presents a fundamental challenge in perception-driven reasoning, where an intelligent system must infer ongoing activities from a vast, unconstrained space of possibilities. Unlike closed-set classification, where predefined categories limit ambiguity, open-world recognition requires models to adaptively infer plausible activities, even when faced with unknown or unseen combinations of actions and objects. This challenge is particularly relevant for assistive robotics, wearable AI, and autonomous agents, where real-time understanding of human actions is crucial for interaction and decision-making. While multimodal foundation models such as Vision-Language Models (VLMs) have demonstrated remarkable zero-shot generalization, their reliance on exhaustive enumeration for inference makes them inefficient for large-scale open-world reasoning. To bridge this gap, structured knowledge is essential to move beyond naive brute-force enumeration, enabling scalable and efficient activity understanding in real-world applications.

\begin{figure*}
    \begin{tabular}{cc}
    \includegraphics[width=0.55\textwidth]{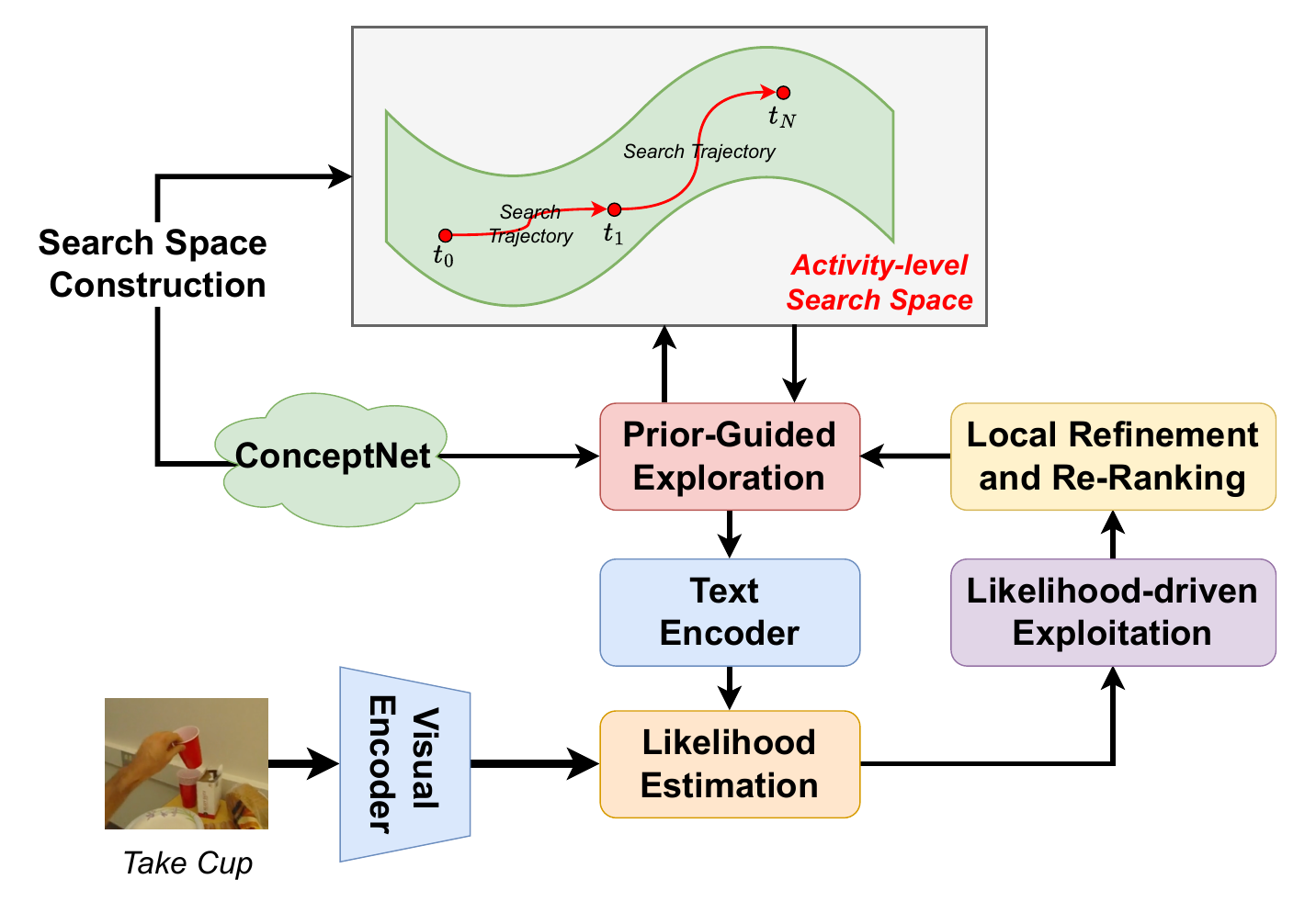} & 
         \includegraphics[width=0.4\textwidth]{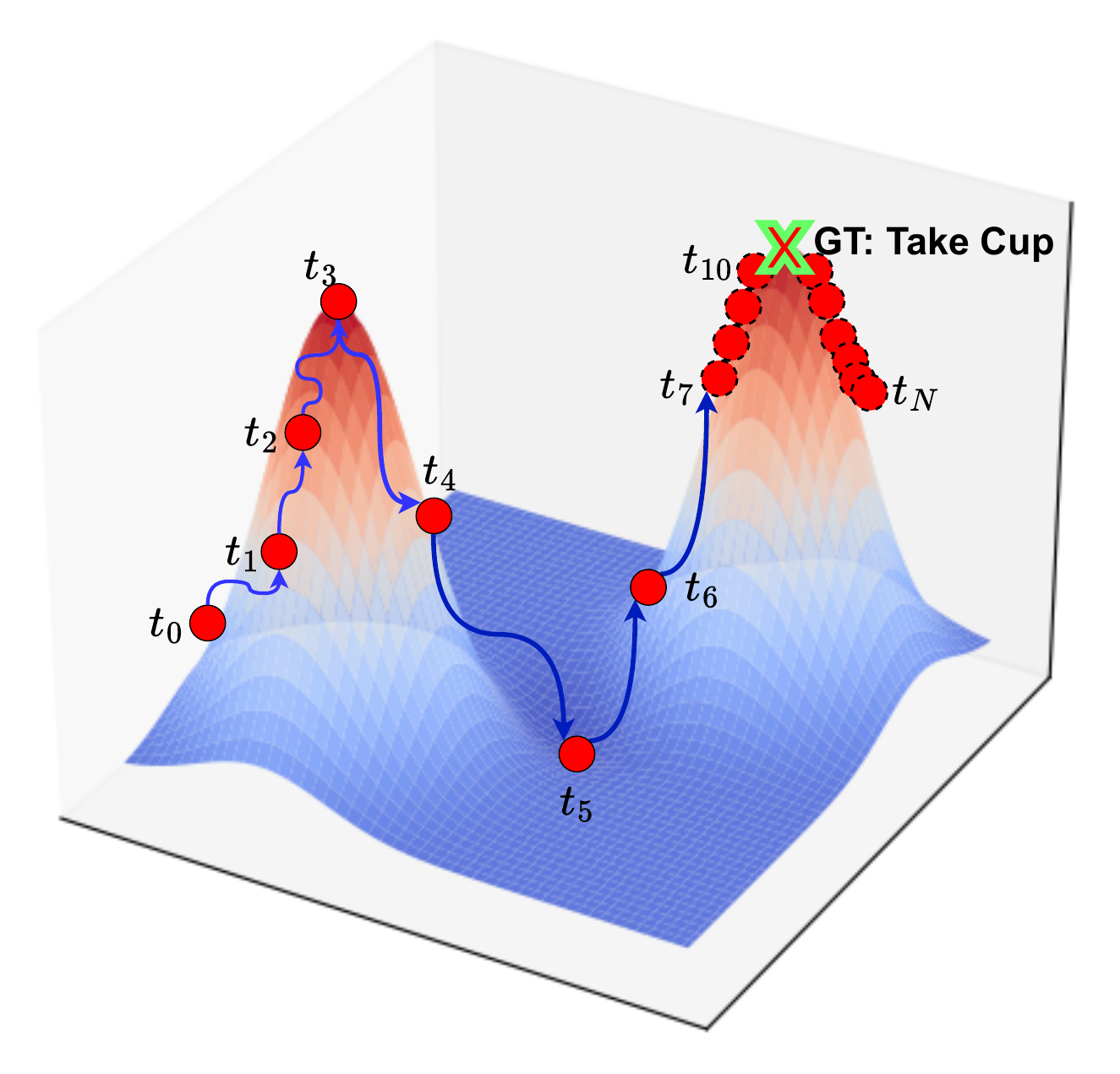} \\
         (a) & (b)\\
    \end{tabular}
    \caption{
    \textbf{(a) ProbRes framework} for open-world egocentric activity recognition. The search space is structured using ConceptNet priors, enabling guided exploration. The model iteratively refines candidates via likelihood estimation, balancing exploration and exploitation, followed by local refinement and re-ranking. (\textbf{b) Search trajectory visualization}, showing how ProbRes navigates likelihood regions to reach high-confidence predictions near the ground truth.
    % (a) \textbf{ProbRes framework for open-world egocentric action recognition.} The search space is structured using commonsense knowledge from ConceptNet, guiding a prior-driven exploration process. The search iteratively refines candidates using a text encoder and likelihood estimation, balancing prior-guided exploration and likelihood-driven exploitation, followed by local refinement and re-ranking for final prediction. (b) \textbf{Illustration of the search trajectory} in the activity search space, where the model navigates through different likelihood regions, adapting search behavior to reach high-confidence predictions near the ground truth.
    }
    \label{fig:arch}
\end{figure*}

The increasing focus on open-world learning~\cite{kundu_discovering_2024,aakur_knowledge_2022,dong_open_2022,gu_open-vocabulary_2022} has led to ambiguity in defining openness, as different factors, objects, actions, and domains can independently contribute to it. Without a clear distinction, comparing methods and assessing generalization remain challenging. To address this, we define a hierarchy of openness levels that systematically capture increasing uncertainty in the search space (Figure~\ref{fig:taxonomy}).
% To systematically study open-world egocentric activity recognition, we define a hierarchy of openness levels, each presenting increasing uncertainty in the search space. 
% Figure~\ref{fig:taxonomy} illustrates this growing complexity and the progressively unconstrained search spaces. 
At \textbf{L0}, as in traditional zero-shot learning, all possible activity categories (action-object combinations) are predefined, but some may lack training examples. This setting benefits directly from Vision-Language Models (VLMs), which can generalize to unseen instances but operate within a bounded conceptual space. At \textbf{L1}, the atomic concept space — objects and actions — is known, but their composition (activities) is undefined, requiring models to reason about plausible pairings. At \textbf{L2}, the domain is known (e.g., cooking), but the underlying search space is unconstrained. Finally, at \textbf{L3}, no prior knowledge about the domain, actions, or objects is available, requiring models to construct the entire search space on the fly. As openness increases, exhaustive enumeration becomes intractable and requires structured priors for adaptive search. 

To address these, we introduce Probabilistic Residual Search (ProbRes), a structured search framework designed to navigate unconstrained activity spaces efficiently. Instead of relying solely on VLM-based likelihood estimation, ProbRes integrates structured priors from external knowledge sources to guide exploration, ensuring semantically meaningful search trajectories without supervision. As illustrated in Figure~\ref{fig:arch}, ProbRes operates in three key phases: (1) exploration, where prior-driven sampling identifies plausible activity candidates; (2) exploitation, where the search refines predictions based on VLM likelihoods; and (3) residual refinement, which further optimizes results by re-ranking activities based on their decomposed action-object components. By balancing structured priors with data-driven reasoning, ProbRes efficiently locates high-likelihood activities while drastically reducing VLM queries.

Our \textbf{contributions} are four-fold: (i) we propose \textit{ProbRes}, a probabilistic residual search strategy that integrates structured priors with likelihood-driven inference, effectively balancing exploration and exploitation for open-world activity recognition, (ii) we introduce a clear hierarchical taxonomy (L0–L3) to systematically define and evaluate open-world recognition challenges, providing a principled foundation for future research, (iii) we show that ProbRes drastically reduces VLM queries while maintaining or improving recognition accuracy, demonstrating the importance of structured priors in large-scale search, and (iv) we highlight the limitations of current VLM-based text embeddings for search efficiency with qualitative analysis and highlight the need for improved semantic structuring.

\section{Related Work}
\textbf{Egocentric video understanding} has gained significant traction in the computer vision community in recent years due to its critical applications in areas such as imitation learning for robotic skills \cite{sermanet2018time}, virtual reality \cite{han2020megatrack}, and autonomous skill acquisition \cite{NEURIPS2022_161c94a5, Lee_2024_CVPR,10.1007/978-3-031-19778-9_38}. A wide range of tasks have been proposed to analyze egocentric video understanding, including procedure understanding, EgoPose \cite{grauman_ego4d_2022, grauman2024ego}, proficiency estimation, question answering \cite{fan_egovqa_2019}, gaze prediction \cite{aakur_unsupervised_2021,li_learning_2013,fathi_learning_2012}, activity recognition \cite{li_deep_2019}, and summarization \cite{lu_story-driven_2013}, driven by the development of large-scale datasets such as EgoExo4D \cite{grauman2024ego}, Ego4D \cite{grauman_ego4d_2022}, EPIC-Kitchens \cite{damen_epic-kitchens_2020}, Charades-Ego \cite{sigurdsson_actor_2018}, GTEA Gaze \cite{li_learning_2013}, GTEA Gaze Plus \cite{fathi_learning_2012}, and Assembly101 \cite{sener2022assembly101}.

Despite these advancements, action recognition in egocentric videos remains challenging, particularly under conditions of rapid camera movement and significant occlusion, leading to issues with accuracy and generalization across different environments. While early approaches to egocentric action recognition were dominated by supervised learning methods, such as spatial-temporal modeling \cite{sudhakaran_lsta_2019}, motion-based representations \cite{ma_going_2016}, hand-object interactions \cite{wang_interactive_2021,zhou_cascaded_2016,xu_egocentric_2025}, and time-series modeling \cite{ryoo_pooled_2015}, recent research has shifted towards self-supervised learning \cite{zhang2023modeling, xue2023learning}, leveraging large-scale pretraining followed by fine-tuning for downstream tasks. Both paradigms remain data-driven and require extensive training to achieve optimal performance. Multi-modal LLM-based models such as MM-Ego \cite{ye_mm-ego_2024} and GPT4Ego \cite{dai_gpt4ego_2024} have recently been explored for egocentric action recognition. Zero-shot learning \cite{sigurdsson_actor_2018,zhang_first-person_2017} offers a potential avenue to reduce dependence on labeled datasets, but its effectiveness is constrained by the inherent limitations of pre-trained models in reasoning about open-world activity compositions.

\textbf{Open-world understanding}, particularly in egocentric activity recognition, remains an underexplored area. While significant progress has been made in open-world object detection, such as Open World DETR \cite{dong_open_2022} and open-vocabulary object detection \cite{gu_open-vocabulary_2022,du_learning_2022}, which leverage VLMs through prompting and knowledge distillation, challenges remain in handling novel class biases, as highlighted by UMB \cite{xi_umb_2024}. One of the first works addressing open-world egocentric activity recognition, KGL \cite{aakur_knowledge_2022}, employed a neuro-symbolic framework utilizing ConceptNet \cite{speer2017conceptnet} for knowledge-driven inference. However, its reliance on object detectors for grounding concepts and the semantic gap between knowledge bases and video data limits its applicability. Chatterjee et al. \cite{chatterjee2023opening} proposed a verb encoder and prompt-based object encoder for open-vocabulary action recognition in egocentric videos. More recently, ALGO \cite{kundu_discovering_2024} extended neuro-symbolic prompting to refine action-object affinities iteratively, combining object grounding with video foundation models. Other works explore related directions, such as incremental open-world action recognition \cite{prijatelj_human_2024} and skeleton-based methods for open-set action recognition \cite{peng_navigating_2023}. 
% Despite these advances, existing approaches struggle in unconstrained search spaces.
 
\textbf{Vision-Language Models} have gained prominence following the success of transformer-based LLMs such as BERT \cite{devlin_bert_2019}, RoBERTa \cite{liu_roberta_2019}, GPT series \cite{radford_improving_2018,radford_language_2019,brown_language_2020,bubeck_sparks_2023}, and open-source models like LLaMA \cite{touvron_llama_2023-1,touvron_llama_2023,grattafiori_llama_2024} and DeepSeek \cite{deepseek-ai_deepseek-r1_2025}. Image-language models, including CLIP \cite{radford_learning_2021}, DeCLIP \cite{li_supervision_2022}, and ALIGN \cite{jia_scaling_2021}, leverage large-scale image-text pairs with contrastive learning \cite{chen_simple_2020,khosla_supervised_2020} to achieve strong zero-shot classification. Video-language models extend these capabilities to egocentric video understanding. EGO-VLP \cite{lin_egocentric_2022} employs separate video and text encoders, Hier-VL \cite{ashutosh_hiervl_2023} captures multi-scale video information hierarchically and LAVILA \cite{zhao_learning_2022} enhances video-text embeddings with LLM-generated narrations. EgoVLPv2 \cite{pramanick_egovlpv2_2023} integrates cross-modal fusion, while ALANAVLM \cite{suglia_alanavlm_2024} advances large-scale multimodal pretraining for embodied AI. However, these models are inherently constrained by the fixed search space defined during inference, limiting their generalization ability in open-world scenarios. 
% In this work, we introduce ProbRes, which dynamically explores and refines activity predictions beyond predefined label sets by integrating structured priors and adaptive search mechanisms.  
% \textbf{Test time scaling} has shown promise in recent times to be more effective than scaling the model parameters during training. \cite{snell_scaling_2024} shows that optimally scaling the test time can be surprisingly more effective than scaling model parameters in terms of LLM performance.\cite{muennighoff_s1_2025} demonstrates improved performance in the code generation benchmark. \cite{li_s_2025} achieves performance improvement with simple test time scaling.  \cite{chen_expanding_2025} shows that jointly scaling model size, training data, and test time compute leads to substantial performance gains in multi-modal vision models. \cite{lin_investigating_2025} suggests that inference time scaling with Multi-modal Chain of Thoughts can improve the reasoning performance of multi-modal models.

% \clearpage
\section{ProbRes: Jump Diffusion-based Reasoning}
\textbf{Overview.} We aim to infer an activity label $a \in \mathcal{S}$ from a given video feature $v$. Unlike conventional closed-set recognition~\cite{sudhakaran_lsta_2019}, where all activities are known during training, open-world settings require reasoning over a vast, partially observed search space, making exhaustive enumeration computationally prohibitive. Given a prior probability distribution $P_{\text{prior}}(a)$ capturing activity occurrence, a Vision-Language Model (VLM) likelihood estimator $P_{\text{likelihood}}(v{\mid}a)$, and a search space $\mathcal{S}$, the goal is to estimate the most probable activity efficiently:

\begin{equation}
    a^* = \arg\max_{a \in \mathcal{S}} P_{\text{likelihood}}(v \mid a).
    \label{ideal_label_eqn}
\end{equation}

However, directly evaluating all candidates in $\mathcal{S}$ is computationally inefficient and often infeasible as the search space grows. Instead, we propose \textit{Probabilistic Residual Search (ProbRes)}, an adaptive search framework that efficiently navigates the text embedding space of activities while being guided by symbolic priors derived from a structured knowledge graph. The overall procedure is summarized in Algorithm~\ref{alg:probres}. ProbRes operates in three key phases: (1) \textit{Exploration}, where search hypotheses are sampled from a prior-driven distribution; (2) \textit{Exploitation}, where search is refined using VLM-based likelihoods; and (3) \textit{Residual Refinement}, where candidates are refined using fine-grained re-ranking mechanism.

\begin{algorithm}[t]
\caption{Probabilistic Residual Search (ProbRes) for Open-World Activity Recognition}
\label{alg:probres}
\textbf{Input}: Prior probabilities $P_{\text{prior}}(a)$, VLM likelihoods $P_{\text{likelihood}}(v \mid a)$, search space $\mathcal{S}$, exploration weight $\lambda$ and duration $T_{\text{explore}}$, max iterations $T$.\\
\textbf{Output}: Predicted activity $a^*$
\begin{algorithmic}[1]
\State \textbf{Initialize:} $t \leftarrow 0$, $a_0 \sim P_{\text{prior}}(a)$ 
% \Comment{\textit{Sample initial activity from prior}}
\While{$t < T$} \Comment{\textit{Iterative Search Process}}
    \If{$t < T_{\text{explore}}$} \Comment{\textit{Exploration Phase}}
        \State $P_{\text{explore}}(a) = \frac{\lambda P_{\text{prior}}(a) + (1-\lambda) \frac{1}{|\mathcal{S}|}}{\sum_{a'} \lambda P_{\text{prior}}(a') + (1-\lambda)}$
        \State $a_{t+1} \sim P_{\text{explore}}(a)$ \Comment{\textit{Sample with exploration}} 
    \Else \Comment{\textit{Exploitation Phase}}
        \State $P_{\text{guided}}(a) \propto P_{\text{prior}}(a)\cdot P_{\text{likelihood}}(v \mid a)$
        \State $a_{t+1} \sim P_{\text{guided}}(a)$ \Comment{\textit{Sample with exploitation}} 
    \EndIf
    \State $t \leftarrow t + 1$
\EndWhile
\State $\mathcal{A}_{\text{refine}} = \text{topk}_{a \in \mathcal{S}} P_{\text{likelihood}}(v \mid a)$ \Comment{\textit{Top $k$ candidates for refinement}}
\For{each $a \in \mathcal{A}_{\text{refine}}$}
    \State $a \rightarrow (a_{\text{action}}, a_{\text{object}})$ \Comment{\textit{Decompose activity}}
    \State $S_a = v^T \phi(a_{\text{action}})$ \Comment{\textit{Action Likelihood}}
    \State $S_o = v^T \phi(a_{\text{object}})$ \Comment{\textit{Object Likelihood}}
    \State $S_{\text{final}}(a) = P_{\text{likelihood}}(v \mid a) + \lambda_a S_a + \lambda_o S_o$
\EndFor
\State \textbf{Return:} $a^* = \arg\max_{a \in \mathcal{A}_{\text{refine}}} S_{\text{final}}(a)$
\end{algorithmic}
\end{algorithm}

\subsection{Constructing and Structuring the Search Space}

We combine knowledge-driven priors with embedding-based organization to efficiently navigate the open-world activity space. The prior space estimates the plausibility of action-object pairs using structured commonsense knowledge, guiding search toward semantically valid activities. In parallel, the search space structuring reorders VLM embeddings to ensure local jumps during search remain meaningful. This structured search space allows ProbRes to make jumps that respect semantic locality while maintaining computational efficiency. 

\subsubsection{Constructing a prior space.} To guide search in open-world activity recognition, we construct a prior probability distribution over action-object pairs using external semantic knowledge. Specifically, we leverage ConceptNet~\cite{speer_conceptnet_2017,speer2017conceptnet} to estimate the likelihood of an action-object pair occurring based on structured commonsense relationships. Given a finite set of actions $\mathcal{A}$ and objects $\mathcal{O}$, we define the prior probability of an activity $a = (a_{\text{action}}, a_{\text{object}})$ as $P_{\text{prior}}(a) \propto f(a_{\text{action}}, a_{\text{object}})$, where $f(a_{\text{action}}, a_{\text{object}})$ is a semantic score computed as the average weight of all shortest paths in ConceptNet between each object and action pair. While this is a good semantic similarity measure, it does not capture the compositional nature of affordance-based affinity between the concepts. 

\textbf{Affinity Scoring.} We model ConceptNet as a directed graph $G$ with nodes as concepts (e.g., \textit{wash}, \textit{dish}) and weighted edges representing semantic relations. The score $f(a_{\text{action}}, a_{\text{object}})$ is derived from the shortest path between $a_{\text{action}}$ and $a_{\text{object}}$, computed as the sum of edge weights with an exponential decay factor $\lambda^i$ at step $i$. To refine the prior, we apply a relation-based adjustment that penalizes negative relations (e.g., \texttt{NotCapableOf}, \texttt{NotUsedFor}) and boosts positive ones (e.g., \texttt{UsedFor}, \texttt{CapableOf}). Specifically, we compute the final affinity score as 
\begin{equation}
    f(a_{\text{action}}, a_{\text{object}}) \leftarrow f(a_{\text{action}}, a_{\text{object}}) \cdot R(a_{\text{action}}, a_{\text{object}})
\end{equation}
where $R(a_{\text{action}}, a_{\text{object}})$ is a multiplicative weight derived from ConceptNet relations. The prior is then normalized across all candidate activities to form a probability distribution, serving as the exploration distribution in ProbRes to guide search toward semantically plausible activities before likelihood-based refinement. Empirically, in Section~\ref{sec:results}, we see that this prior space guides the search process much more efficiently than a random search. 

\subsubsection{Search Space Structuring}
The search is performed in the text embedding space of activities, where each candidate is represented by a Vision-Language Model (VLM) embedding $\phi(a)$. However, raw VLM embeddings do not always encode the fine-grained affordance relationships necessary for effective search. We construct an ordered search space to ensure that the search moves in semantically meaningful directions. First, we extract action-object phrase embeddings from the VLM to form an initial unordered set $\mathcal{S} = \{a_i = (a_{\text{action}}, a_{\text{object}})\}$. We define a similarity metric based on the Euclidean distance between embeddings, $d(a_i, a_j) = ||\phi(a_i) - \phi(a_j)||$, where $\phi(a)$ represents the embedding of activity $a$. The search space is structured by selecting an anchor point $a_{\text{ref}}$ and sorting activities based on their distance to $a_{\text{ref}}$. This ensures that similar activities remain proximate, enabling ProbRes to perform jumps that respect semantic locality.

\subsection{Adaptive Search via Jump Diffusion}
Probabilistic Residual Search (ProbRes) efficiently estimates the most probable activity from a large open-world search space without exhaustive evaluation. It iteratively refines predictions by optimizing for

% \begin{equation}
%     a^* = \arg\max_{a \in \mathcal{S}} \left[ P_{\text{likelihood}}(v \mid a) + \lambda_a S_a + \lambda_o S_o \right]
%     \label{eq:search_optimization}
% \end{equation}
\begin{equation}
    a^* = \arg\max_{a \in \mathcal{S}} \left[ P_{\text{search}}(a) + \lambda_a S_a + \lambda_o S_o \right]
    \label{eq:search_optimization}
\end{equation}
where $P_{\text{search}}(a) = P_{\text{explore}}(a)$ for $t < T_{\text{explore}}$ (exploration phase, Eqn~\ref{eqn:explore}) and transitions to $P_{\text{search}}(a) = P_{\text{guided}}(a)$ for $t \geq T_{\text{explore}}$ (exploitation phase, Eqn.~\ref{eqn:exploit}). $P_{\text{explore}}(a)$ leverages prior knowledge to guide broad search, while $P_{\text{guided}}(a)$ refines the search trajectory using likelihood estimates. $S_a$ and $S_o$ capture the fine-grained likelihood of the activity’s action and object components. A local refinement step does a fine-grained search in high-likelihood areas to distill activity labels. We detail this process next.
% Finally, a concept decomposition step re-ranks the top candidates. 
% To this end, ProbRes follows a three-stage search: (1) Prior-guided exploration and likelihood-based exploitation to locate high-probability regions in the embedding space, (2) Localized refinement within these regions, and (3) Concept decomposition to verify and re-rank the top candidates independently. 

\subsubsection{Sampling-based Search}
A key challenge in search-based optimization is efficiently identifying high-likelihood regions in the large search space while avoiding excessive VLM consultations, which can be expensive. We address this by balancing exploration (prior-guided sampling to discover diverse candidates) and exploitation (likelihood-guided refinement to focus on promising hypotheses). This two-phase process ensures the search remains computationally efficient while converging to high-likelihood solutions. 

\textbf{Exploration as Prior-Guided Search.} 
In the initial phase, the search process favors broad exploration to efficiently locate high-likelihood regions in $\mathcal{S}$. Instead of randomly sampling from the entire space, ProbRes leverages prior knowledge to guide exploration, ensuring that early search iterations prioritize semantically plausible activities:

\begin{equation}
    P_{\text{explore}}(a) = \frac{\lambda P_{\text{prior}}(a) + (1-\lambda) \frac{1}{|\mathcal{S}|}}{\sum_{a'} P_{\text{prior}}(a') + (1-\lambda)}
    \label{eqn:explore}
\end{equation}

Here, $\lambda \in [0,1]$ controls the tradeoff between prior-driven and uniform random sampling. When $\lambda \approx 1$, the search is strongly guided by priors, favoring activities that align with commonsense affordances. When $\lambda \approx 0$, the search behaves like a random walk over $\mathcal{S}$. This mechanism allows the search to avoid implausible candidates.

\textbf{Exploitation as Likelihood-Guided Refinement}. 
Once high-likelihood regions begin to emerge, the search shifts focus to exploitation. The goal is to refine the search trajectory and concentrate computational resources on the most promising candidates. Instead of sampling from the entire prior space, the search becomes likelihood-driven $P_{\text{guided}}(a) \propto P_{\text{likelihood}}(v \mid a)$. Hypotheses are then selected based on their estimated likelihood given the observed video, prioritizing candidates that the VLM deems most probable. This ensures the search process refines the results rather than continuing broad exploration.

% \textbf{Balancing Exploration and Exploitation.} The transition from exploration to exploitation is governed by a temperature parameter $\tau$, which gradually decreases over time 
% \begin{equation}
%     P_{\text{guided}}(a) = \frac{\exp(P_{\text{likelihood}}(v \mid a) / \tau)}{\sum_{a'} \exp(P_{\text{likelihood}}(v \mid a') / \tau)}.
%     \label{eqn:exploit}
% \end{equation}
% Early in the search, a high $\tau$ value allows for diverse exploration, while a low $\tau$ value later in the process ensures focused refinement. 
% This annealing strategy efficiently balances discovering new hypotheses and converging on the best solution. 
% The search initially favors exploration before gradually transitioning to exploitation, driven by integrating priors and likelihood estimates, where priors dominate early in the search and likelihoods take precedence as more information becomes available. 
% This formulation helps identify high-likelihood activities while requiring only a small fraction of the computational budget of an exhaustive search. 

\textbf{Balancing Exploration and Exploitation.} The transition from exploration to exploitation is governed by integrating prior probabilities and likelihood estimates. Instead of relying on an explicit temperature parameter $\tau$, the search gradually shifts towards exploitation by weighting priors and likelihoods dynamically:
\begin{equation}
    P_{\text{guided}}(a) = \frac{P_{\text{prior}}(a) P_{\text{likelihood}}(v \mid a)}{\sum_{a'} P_{\text{prior}}(a') P_{\text{likelihood}}(v \mid a')}
    \label{eqn:exploit}
\end{equation}
Early in the search, priors provide a structured means of exploration, ensuring that plausible activities are prioritized. As the search progresses and likelihood estimates become more informative, the search increasingly favors high-likelihood candidates while retaining prior knowledge for regularization. This gradual transition enables an efficient balance between discovering new hypotheses, refining the most probable ones, and reducing computational overhead.

\begin{table*}[ht]
    \centering
    \setlength{\tabcolsep}{2pt} % Reduce column spacing
    \renewcommand{\arraystretch}{1.2} % Adjust row spacing
    \resizebox{\textwidth}{!}{%
    \begin{tabular}{|c|cccc|cccc|cccc|cccc|}
        \toprule[1.2pt]
        \vspace{-3pt}
        \multirow{3}{*}{\textbf{Approach}} & \multicolumn{16}{c|}{\textbf{Datasets}} \\
        \cmidrule(lr){2-17}
        & \multicolumn{4}{c|}{\textbf{GTEA Gaze}} & \multicolumn{4}{c|}{\textbf{GTEA Gaze+}} & \multicolumn{4}{c|}{\textbf{EK100}} & \multicolumn{4}{c|}{\textbf{CharadesEgo}} \\
        \cmidrule(lr){2-17}
        & \textbf{VLM Calls} & \textbf{Activity} & \textbf{Phrase} & \textbf{WUPS} & \textbf{VLM Calls} & \textbf{Activity} & \textbf{Phrase} & \textbf{WUPS} & \textbf{VLM Calls} & \textbf{Activity} & \textbf{Phrase} & \textbf{WUPS} & \textbf{VLM Calls} & \textbf{Activity} & \textbf{Phrase} & \textbf{WUPS} \\
        \midrule[1.2pt]
        ALGO & N/A* & 15.06 & 1.80 & 46.89 & N/A* & 18.84 & 3.40 & 43.29 & N/A* & 8.49 & 1.72 & 38.53 & N/A* & 2.62 & 0.10 & 36.68\\
        KGL & N/A** & 6.58 & 0.30 & 35.83 & N/A** & 10.76 & 1.10 & 39.41 & N/A** & 3.07 & 0.94 & 36.85 & N/A** & 2.06 & 0.05 & 31.48\\
        ALGO+LAVILA & N/A* & 22.05 & 3.50 & 49.42 & N/A* & 28.87 & 8.30 & 53.38 & N/A* & 17.69 & 4.39 & 34.47 & N/A* & 2.94 & 0.21 & 37.21\\
        LAVILA-Decomp & 380 & 14.57 & 2.10 & 48.65 & 405 & 26.87 & 7.89 & 53.12 & 29100 & 17.21 & 4.28 & 43.37 & 1254 & 11.82 & 1.51 & 38.99\\
        EGOVLP-Decomp & 380 & 9.31 & 0.90 & 40.51 & 405 & 23.30 & 4.10 & 51.74 & 29100 & 15.14 & 2.05 & 39.94 & 1254 & 10.66 & 1.12 & 38.73\\
        LAVILA & 380 & 29.02 & 9.53 & 51.31 & 405 & 29.82 & 7.91 & 53.27 & 29100 & 18.81 & 4.43 & 43.52 & 1254 & 11.34 & 1.45 & 39.76\\
        EGOVLP & 380 & 17.07 & 1.80 & 46.77 & 405 & 27.00 & 3.95 & 51.48 & 29100 & 12.78 & 2.22 & 39.84 & 1254 & 10.55 & 1.21 & 38.43\\
        \midrule[1.2pt]
        \rowcolor{gray!20} LAVILA + ProbRes (Ours) & 110 & \textbf{33.80} \increase{} & \textbf{8.98} \increase{} & \textbf{53.34} \increase{} & 175 & \textbf{31.70} \increase{} & \textbf{9.33} \increase{} & \textbf{53.82} \increase{} & 3000 & \textbf{18.89} \increase{} & \textbf{4.61} \increase{} & \textbf{43.55} \increase{} & 300 & \textbf{13.71} \increase{} & \textbf{2.12} \increase{} & \textbf{40.91} \increase{}\\
        \rowcolor{gray!20} EGOVLP + ProbRes (Ours) & 110 & \textbf{19.12} \increase{} & \textbf{1.97} \increase{} & \textbf{49.25} \increase{} & 178 & \textbf{29.84} \increase{} & \textbf{6.32} \increase{} & \textbf{53.98} \increase{} & 3000 & \textbf{13.45} \increase{} & \textbf{2.36} \increase{} & \textbf{40.53} \increase{} & 300 & \textbf{12.65} \increase{} & \textbf{1.78} \increase{} & \textbf{38.91} \increase{}\\
        \bottomrule[1.2pt]
    \end{tabular}%
    }
    \caption{
    \textbf{L1 evaluation results} on GTEA Gaze, GTEA Gaze+, EK100, and CharadesEgo. ProbRes consistently improves activity recognition across all datasets while significantly reducing VLM calls, demonstrating its efficiency in structured search. Green arrows indicate performance gains, while red arrows denote declines. *Frame-wise CLIP querying, **Frame-wise Faster R-CNN querying.
    % L1 evaluation results on GTEA Gaze, GTEA Gaze+, EK100, and CharadesEgo. Green arrows denote an increase; red arrows denote a decrease. *Indicates frame-wise CLIP querying, **indicates frame-wise Faster R-CNN-querying.
    }
    \label{tab:L1_results}
\end{table*}

\subsubsection{Localized Refinement in High-Likelihood Areas}

While the exploitation phase prioritizes high-likelihood candidates, it remains probabilistic and may fail to leverage strong hypotheses fully. ProbRes performs localized refinement to improve robustness, a deterministic re-evaluation of the highest-confidence candidates. Instead of continuing broad sampling, search is restricted to a subset of top-ranked hypotheses ($|\mathcal{A}_{\text{refine}}| = K$) defined as $\mathcal{A}_{\text{refine}} = \arg\max_{a \in \mathcal{S}} P_{\text{likelihood}}(v \mid a)$. Unlike exploitation, which still samples from a distribution, this refinement stage deterministically focuses on fine-tuning a small set of high-confidence candidates. This is crucial because early likelihood estimates can be noisy, and a single-pass sampling strategy may not be sufficient to distinguish the most accurate activity labels. By revisiting the strongest hypotheses, ProbRes prevents premature convergence to suboptimal solutions. 
This localized refinement stabilizes the ranking of predictions, ensuring that minor variations in likelihood do not disrupt the final decision. Instead of additional broad search iterations, the final stage is spent refining existing strong candidates to enhance efficiency and robustness. 

\subsubsection{Concept Decomposition and Re-Ranking}

After obtaining a refined candidate set, ProbRes applies a final re-ranking step, decomposing each activity into its action and object components. Instead of treating activities as atomic units, this step evaluates their semantic consistency separately by computing individual similarity scores for the action and object. 
Concept decomposition mitigates errors from Vision-Language Models (VLMs) by separately evaluating the action and object components, reducing the impact of incorrect phrase-level embeddings. Using finer-grained representations can help compensate for VLM ambiguities and ensure that activities are ranked based on semantically consistent elements rather than potentially noisy joint embeddings. 
Given a video representation $v$ and VLM embeddings $\phi(a_{\text{action}})$ and $\phi(a_{\text{object}})$, we compute their respective alignment scores as $S_a = v^T \phi(a_{\text{action}})$ and $S_o = v^T \phi(a_{\text{object}})$. These scores adjust the likelihood-based ranking by incorporating finer-grained semantic relevance. The final score for re-ranking is defined as $S_{\text{final}} = P_{\text{likelihood}}(v \mid a) + \lambda_a S_a + \lambda_o S_o$, where $\lambda_a$ and $\lambda_o$ control the relative influence of the action and object alignment scores. The highest-ranked candidate from the refined set $\mathcal{A}_{\text{refine}}$ is then selected as $a^* = \arg\max_{a \in \mathcal{A}_{\text{refine}}} S_{\text{final}}$. This hierarchical re-ranking ensures that activities with semantically consistent components receive higher confidence while mitigating errors from overly generic or ambiguous predictions.

\textbf{Implementation Details.} We use EGOVLP~\cite{lin_egocentric_2022} and LAVILA~\cite{zhao_learning_2022} as VLM backbones for likelihood estimation, guided by structured priors from ConceptNet. For L2 and L3 evaluations, search spaces were generated using Gemini 2.0 Flash~\cite{pichai2024gemini2} and refined to remove duplicates. The search process is regulated by the exploration-exploitation balance parameter ($\lambda$) and search iterations ($T$), tuned via grid search for efficiency and accuracy. We set $\lambda{=}0.5$ for prior-guided exploration before likelihood-driven exploitation and cap iterations at $T{=}3000$ for smaller datasets and $T{=}1000$ for larger ones to control computational cost. Re-ranking weights $\lambda_a$ and $\lambda_o$ are optimized within $[0.3, 0.7]$ to balance action-object contributions. All experiments run on an NVIDIA RTX 3090 GPU, with an average inference time of 2 seconds per video.

\section{Experimental Setup}
We evaluate ProbRes on four egocentric activity recognition datasets - GTEA Gaze~\cite{li_learning_2013}, GTEA Gaze+~\cite{fathi_learning_2012}, EPIC-Kitchens-100 (EK100)~\cite{damen_epic-kitchens_2020}, and Charades-Ego~\cite{sigurdsson_actor_2018} - covering structured kitchen tasks to unconstrained daily activities. GTEA Gaze and GTEA Gaze+ feature meal preparation with generic, often ambiguous action-object pairs, making disambiguation challenging due to frequent interaction overlap. EK100 provides a fine-grained, structured activity space with precise action-object interactions, serving as a benchmark for large-scale open-world inference. Charades-Ego, spanning daily activities, requires robust reasoning over diverse activity compositions, further testing generalization in open-world settings.
% These datasets allow us to systematically analyze the effectiveness of ProbRes in navigating different openness levels. 
% By leveraging these benchmarks, we demonstrate the scalability and adaptability of our approach in both domain-specific and general-purpose egocentric activity recognition tasks.

\begin{table*}[ht]
    \centering
    \setlength{\tabcolsep}{5pt} % Adjust column spacing
    \renewcommand{\arraystretch}{1.2} % Adjust row spacing
    \resizebox{\textwidth}{!}{%
    \begin{tabular}{|c|c|c|ccc|ccc|ccc|}
        \toprule
        \multirow{2}{*}{\textbf{Openness}} & \multirow{2}{*}{\textbf{Approach}} & \multirow{2}{*}{\textbf{\# VLM Calls}} & \multicolumn{3}{c|}{\textbf{GTEA Gaze}} & \multicolumn{3}{c|}{\textbf{GTEA Gaze+}} & \multicolumn{3}{c|}{\textbf{EK100}} \\
        \cmidrule(lr){4-12}
        & & & \textbf{Object} & \textbf{Action} & \textbf{Activity} & \textbf{Object} & \textbf{Action} & \textbf{Activity} & \textbf{Object} & \textbf{Action} & \textbf{Activity} \\
        \midrule
        \multirow{4}{*}{\textbf{L2}} 
        & LAVILA & 37191 & {50.72} & 25.96 & 38.34 & 59.25 & \textbf{30.36} & 44.81 & 37.89 & \textbf{23.39} & 30.64 \\
         & \cellcolor{gray!20} LAVILA + ProbRes (Ours) & \cellcolor{gray!20} 1500 & \cellcolor{gray!20} \textbf{53.49}\increase{} & \cellcolor{gray!20} \textbf{33.07}\increase{} & \cellcolor{gray!20} \textbf{43.28}\increase{} & \cellcolor{gray!20} \textbf{62.26}\increase{} & \cellcolor{gray!20} 28.03\decrease{} & \cellcolor{gray!20} \textbf{45.15}\increase{} & \cellcolor{gray!20} \textbf{41.28}\increase{} & \cellcolor{gray!20} 22.56 \decrease{} & \cellcolor{gray!20} \textbf{31.92}\increase{} \\
         \cmidrule(lr){2-12}
        & EGOVLP & 37191 & \textbf{56.77} & 18.34 & 37.56 & 60.77 & 22.27 & 41.52 & 40.87 & \textbf{21.32} & 31.10 \\
        & \cellcolor{gray!20} EGOVLP + ProbRes (Ours) & \cellcolor{gray!20} 1500 & \cellcolor{gray!20} 56.27\decrease{} & \cellcolor{gray!20} \textbf{19.98}\increase{} & \cellcolor{gray!20} \textbf{38.13}\increase{} & \cellcolor{gray!20} \textbf{62.89\increase{}} & \cellcolor{gray!20} \textbf{23.89}\increase{} & \cellcolor{gray!20} \textbf{43.39}\increase{} & \cellcolor{gray!20} \textbf{42.54}\increase{} & \cellcolor{gray!20} 21.07\decrease{} & \cellcolor{gray!20} \textbf{31.81}\increase{} \\
        \midrule
        \multirow{4}{*}{\textbf{L3}} 
        & LAVILA & 195714 & 59.85 & 32.27 & 46.06 & 60.43 & 28.73 & 44.58 & \textbf{41.60} & 20.51 & 31.06\\

        & \cellcolor{gray!20} LAVILA+ProbRes (ours) & \cellcolor{gray!20} 5000 & \cellcolor{gray!20} \textbf{59.89}\increase{} & \cellcolor{gray!20} \textbf{35.53}\increase{} & \cellcolor{gray!20} \textbf{47.71}\increase{} & \cellcolor{gray!20} \textbf{62.35}\increase{} & \cellcolor{gray!20} \textbf{29.04}\increase{} & \cellcolor{gray!20} \textbf{45.70}\increase{} & \cellcolor{gray!20} {41.20}\decrease{} & \cellcolor{gray!20} \textbf{23.96}\increase{} & \cellcolor{gray!20} \textbf{32.58}\increase{}\\
        
        & EGOVLP & 195714 & \textbf{52.94} & 18.93 & 35.94 & 56.95 & 24.37 & 40.66 & \textbf{40.70} & 20.42 & 30.56 \\
        
        & \cellcolor{gray!20} EGOVLP+ProbRes (Ours) & \cellcolor{gray!20} 5000 & \cellcolor{gray!20} {52.01}\decrease{} & \cellcolor{gray!20} \textbf{20.92}\increase{} & \cellcolor{gray!20} \textbf{36.47}\increase{} & \cellcolor{gray!20} \textbf{60.08}\increase{} & \cellcolor{gray!20} \textbf{24.75}\increase{} & \cellcolor{gray!20} \textbf{42.42}\increase{} & \cellcolor{gray!20} {39.91}\decrease{} & \cellcolor{gray!20} \textbf{24.13}\increase{} & \cellcolor{gray!20} \textbf{32.02}\increase{} \\
        \bottomrule
    \end{tabular}%
    }
    \caption{
    \textbf{L2 and L3 evaluation} results on GTEA Gaze, GTEA Gaze+, and EK100. ProbRes achieves significant accuracy gains while reducing VLM calls, demonstrating efficient search in unconstrained spaces. WUPS is reported for object, action, and activity levels.
    % Green arrows indicate improvements, while red arrows denote declines.
    % L2 and L3 evaluation results on GTEA Gaze, GTEA Gaze+, and EK100. Each dataset includes results for Object, Action, and Activity using the WUPS metric.
    }
    \label{tab:L2_L3_results}
\end{table*}
\textbf{Metrics.} To evaluate performance across different levels of openness, we adopt a suite of metrics tailored to open-world activity recognition. Traditional accuracy and mean Average Precision (mAP) are limited in open-world settings, assuming a closed, fully enumerated label space. Instead, we use WUPS (Wu-Palmer Similarity)~\cite{wu1994verb} to quantify semantic similarity between predicted and ground-truth activity labels, capturing partial correctness even when exact matches are infeasible. WUPS is computed for objects, actions, and full activity labels. We also report exact phrase-level match accuracy to measure precise label retrieval. 

% At the highest openness levels, where activity labels are entirely unconstrained, we complement WUPS with exact match accuracy, which, despite its strictness, provides insight into the model's ability to retrieve precise labels. 

\textbf{Baselines.} We compare ProbRes against both Vision-Language Models (VLMs) and neuro-symbolic approaches to comprehensively evaluate open-world activity recognition. EGOVLP~\cite{lin_egocentric_2022} and LAVILA~\cite{zhao_learning_2022} are strong baselines, leveraging large-scale VLM training to match video embeddings with pre-defined activity labels. While effective in closed-set settings, their reliance on direct embedding similarity limits their adaptability in unconstrained search spaces. To assess the role of structured knowledge, we also evaluate against KGL~\cite{aakur_knowledge_2022} and ALGO~\cite{kundu_discovering_2024}, two neuro-symbolic models that integrate commonsense reasoning for activity inference. KGL utilizes ConceptNet~\cite{speer_conceptnet_2017,speer2017conceptnet} relations to construct knowledge graphs for action-object reasoning, while ALGO refines activity representations through affordance-based priors. 
% However, both methods depend on predefined knowledge structures, making them unsuitable for L2 and L3 evaluation, where the action and object spaces extend beyond known categories. 
% In contrast, ProbRes dynamically explores and refines activity predictions, making it well-suited for open-world inference across varying levels of search space complexity.

\section{Quantitative Results}\label{sec:results}

\textbf{L1 evaluation} results are summarized in Table~\ref{tab:L1_results}. ProbRes achieves higher accuracy with fewer VLM calls, highlighting the efficiency of a structured search over exhaustive querying. On GTEA Gaze, ProbRes reduces VLM queries from $380$ to just $110$, while on EK100, the reduction is even more pronounced—from $29,100$ to just $3,000$, while achieving higher WUPS and phrase-level accuracy across datasets. This efficiency-accuracy tradeoff is critical for real-world scalability, demonstrating that ProbRes can effectively navigate open-world spaces without brute-force enumeration. The performance gains hold even in larger, more structured datasets like EK100 and CharadesEgo, suggesting that ProbRes generalizes beyond small-scale benchmarks and is well-suited for real-world, large-scale, open-world activity recognition. Interestingly, neuro-symbolic approaches like KGL and ALGO struggle significantly, with KGL performing worse than even baseline VLMs (e.g., $35.83$ WUPS on GTEA Gaze). This highlights a fundamental limitation of fixed knowledge graphs, which lack adaptability in dynamic, open-ended environments where the space of possible actions is fluid and context-dependent. Furthermore, ProbRes shows significant improvements at the phrase level, indicating that it predicts plausible activities and refines them into more semantically accurate action-object descriptions. 
These findings suggest that leveraging structured priors becomes even more critical for guiding inference as the search space complexity increases exponentially in L2 and L3 settings. The ability of ProbRes to reduce VLM reliance helps scale to unconstrained activity recognition tasks. 

\textbf{L2 Evaluation} results are presented in Table~\ref{tab:L2_L3_results}. In this setting, the search space is constructed dynamically by querying an LLM for domain-specific (e.g., kitchen) actions and objects (see supplementary for details). This leads to a significantly expanded and fine-grained space, introducing high specificity (e.g., \textit{julienne}, \textit{emulsify}, \textit{ladle}) and semantic overlap (e.g., \textit{stir} vs. \textit{whisk}), increasing the challenge of accurate retrieval. Despite this complexity, ProbRes demonstrates strong performance while drastically reducing VLM calls—from $37,191$ to just $1,500$, maintaining or improving WUPS across datasets. Notably, while ProbRes improves activity recognition scores overall, it particularly benefits action recognition, which is expected given the disambiguation challenges posed by the expanded action space. Interestingly, EGOVLP sees a small drop in object-level WUPS when paired with ProbRes, suggesting that its lower baseline object retrieval performance may limit downstream phrase and activity accuracy. 
These results reinforce our hypothesis that structured priors and guided search enable efficient inference in unconstrained spaces to balance specificity and generalization.

\begin{figure*}
    \centering
    \begin{tabular}{ccc}
         \includegraphics[width=0.3\linewidth]{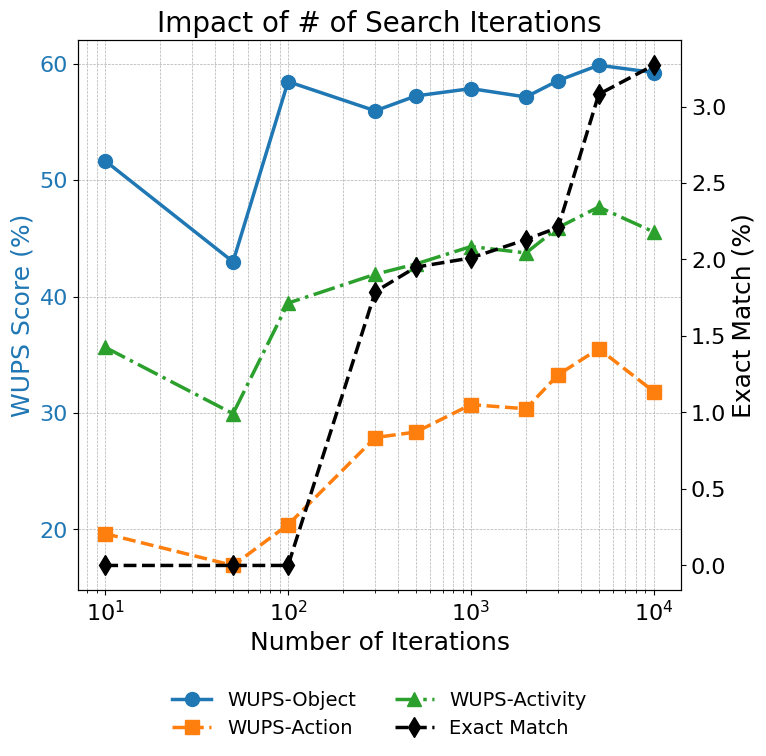} & 
         \includegraphics[width=0.3\linewidth]{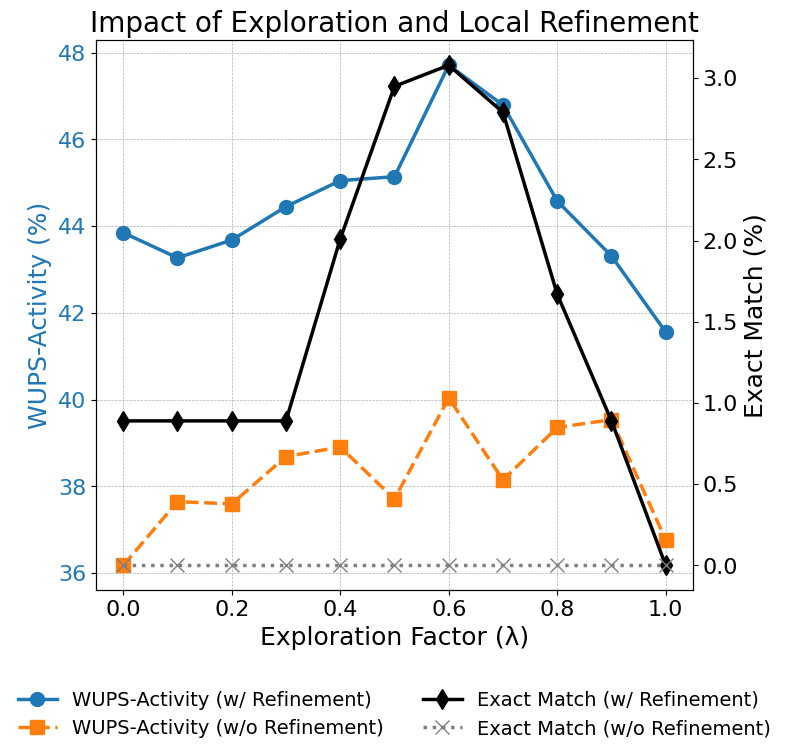} & 
         \includegraphics[width=0.3\linewidth]{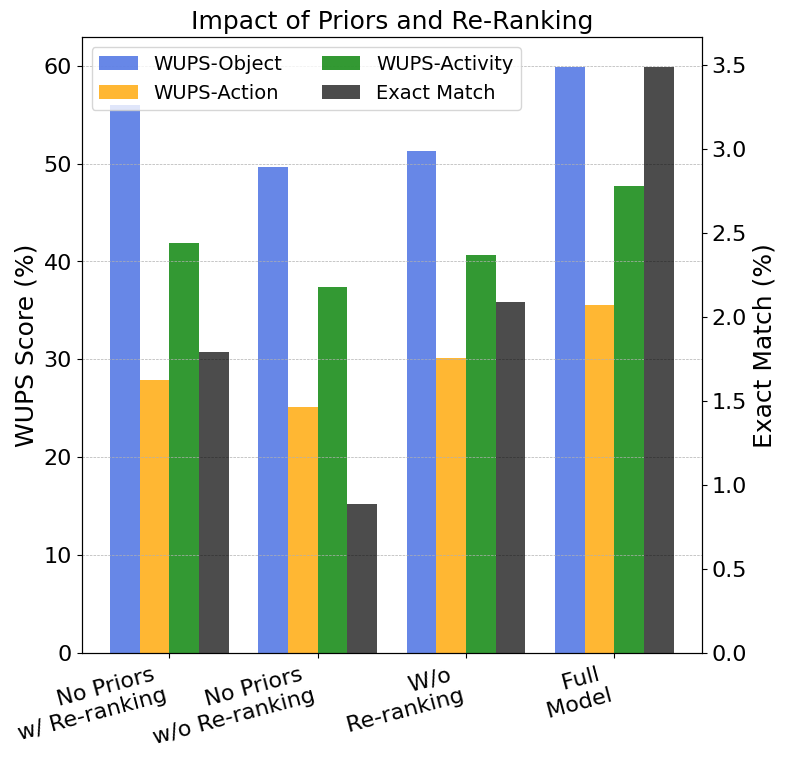} \\
         (a) & (b) & (c)\\
    \end{tabular}
    \caption{\textbf{Ablation Studies} that illustrate the impact of  (a) the number of search iterations, (b) exploration-exploitation tradeoff and local refinement, and (c) knowledge-based priors and concept decomposition-based re-ranking, on the final performance.}
    \label{fig:ablation}
\end{figure*}

\textbf{L3 Evaluation} introduces a significantly more unconstrained setting where even the domain is unknown, necessitating the automatic construction of the search space. We use an LLM to generate a broad set of everyday actions and objects spanning multiple domains. 
(More details in the supplementary materials.) 
Unlike L2, which constrained search to kitchen-related activities, L3 contains highly generic activities such as "carry chair", "store container", or "hold book", alongside more specific ones like "boil pasta" or "slice onion". This results in a vastly expanded search space with increased ambiguity, making exhaustive enumeration impractical. Interestingly, in many cases, most models perform better in L3 than in L2. 
This is primarily due to the LLM-generated search space being more generic, allowing for better coverage of ground-truth activities. However, this also highlights a key insight: \textit{the success of open-world recognition depends on the quality of the search space construction.} An uncurated or overly broad search space may introduce excessive noise, while an excessively restrictive one may fail to capture plausible activities. As shown in Table~\ref{tab:L2_L3_results}, ProbRes outperforms baselines despite the less semantically structured search space. The reliance on structured priors is even more pronounced here, as VLMs struggle with the broad combinatorial possibilities, and brute-force methods quickly become infeasible. Notably, the performance gap between ProbRes and naive VLM inference increases in L3, reinforcing the need for an intelligent search strategy in open-ended environments. These findings suggest that scalable open-world activity recognition requires methods capable of dynamically adapting the search space, making ProbRes a promising direction for real-world deployment.

\textbf{Ablation Studies.} We analyze the impact of key ProbRes components on open-world activity recognition using LAVILA as the VLM backbone on GTEA Gaze under L3 evaluation (Figure~\ref{fig:ablation}). 
Figure~\ref{fig:ablation}(a) examines search iterations. While WUPS scores steadily improve, exact match plateaus, highlighting diminishing returns from over-exploration. This tradeoff is more evident in larger datasets like EK100, where broader search improves semantic accuracy but reduces exact match likelihood. Figure~\ref{fig:ablation}(b) evaluates the exploration-exploitation tradeoff by varying $\lambda$. Pure likelihood-driven search ($\lambda = 0$) leads to premature convergence on high-confidence but incorrect activities, while excessive prior-driven exploration degrades performance. Optimal balance ($\lambda \approx 0.6$) yields the highest WUPS and exact match, emphasizing the need for guided search. Figure~\ref{fig:ablation}(c) ablates knowledge priors and re-ranking. Removing ConceptNet priors significantly reduces performance, particularly in L3, where an unstructured search space increases uncertainty. Eliminating re-ranking lowers exact match, confirming its role in refining semantically similar but incorrect predictions. 
% The full ProbRes model—integrating priors, guided search, and refinement—achieves the best results, demonstrating its effectiveness in structured open-world inference.

\begin{figure}[t]
    \centering
    \includegraphics[width=0.99\linewidth]{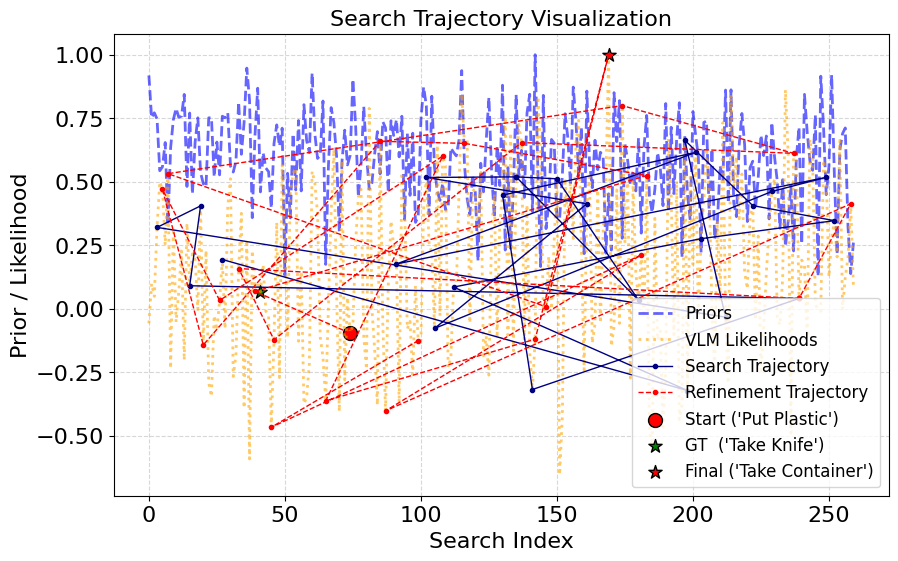}
    \caption{\textbf{Qualitative Visualization} of the search trajectory by ProbRes across different phases indicating exploration,  exploitation, and the final refinement phase.}
    \label{fig:Trejectory_Vis}
\end{figure}

\textbf{Qualitative Analysis.} Figure~\ref{fig:Trejectory_Vis} visualizes the search trajectory of ProbRes in identifying the ground truth (``take fork'') through exploration, exploitation, and refinement. In the exploration phase, the model samples diverse but loosely related actions (e.g., ``put plastic'' and ``cut freezer''), demonstrating the influence of structured priors in steering broad yet relevant candidates. However, the unstructured nature of VLM embeddings results in abrupt transitions between unrelated activities (e.g., ``pour burner'' to ``compress microwave''), disrupting semantic coherence. During exploitation, the search progressively narrows to more plausible candidates (e.g., ``take fridge'' and ``put milk''), leveraging likelihood estimates to refine predictions. The refinement phase further structures the trajectory, filtering semantically similar but incorrect activities before converging near the ground truth. The inconsistencies in VLM embedding distances hinder an optimal search path, reinforcing the need for more structured representations~\cite{kundu2023ggt}, such as hyperbolic embeddings~\cite{pal2024compositional}, to improve search efficiency.% in open-world settings.

\section{Limitations, Discussion, and Future Work}

While ProbRes significantly improves open-world egocentric activity recognition through structured search, challenges remain. Its reliance on Vision-Language Models (VLMs) inherits biases from pretraining data, occasionally leading to inefficient search trajectories. Structured priors help mitigate this, but the lack of semantic organization in text embeddings results in erratic jumps, suggesting that hyperbolic embeddings could enforce better locality. In higher openness levels (L2 and L3), LLM-generated search spaces enable scalable inference but require careful curation to avoid overly broad or noisy label distributions. Despite reducing VLM calls, further efficiency improvements—such as hierarchical search or adaptive refinement—are necessary for real-world deployment. ProbRes demonstrates that structured reasoning can replace brute-force enumeration in open-world settings. Integrating commonsense priors with probabilistic search lays the foundation for scalable and interpretable egocentric AI. Future work will explore additional cues, including human-object interactions and scene context, to enhance inference in unconstrained.

\textbf{Acknowledgement.} This research was supported in part by the US National Science Foundation grants IIS 2348689 and IIS 2348690. We thank Dr. Anuj Srivastava (FSU), Dr. Sudeep Sarkar (USF), and the anonymous reviewers for their thoughtful feedback to help present this work better.

% \input{sec/1_intro}
% \input{sec/2_formatting}
% \input{sec/3_finalcopy}
% \clearpage
{
    \small
    \bibliographystyle{ieeenat_fullname}
    \bibliography{main}
}

\end{document}

%% file: sec/0_abstract.tex
\begin{abstract}
Open-world egocentric activity recognition poses a fundamental challenge due to its unconstrained nature, requiring models to infer unseen activities from an expansive, partially observed search space. We introduce ProbRes, a Probabilistic Residual search framework based on jump-diffusion that efficiently navigates this space by balancing prior-guided exploration with likelihood-driven exploitation. Our approach integrates structured commonsense priors to construct a semantically coherent search space, adaptively refines predictions using Vision-Language Models (VLMs) and employs a stochastic search mechanism to locate high-likelihood activity labels while minimizing exhaustive enumeration efficiently. We systematically evaluate ProbRes across multiple openness levels (L0–L3), demonstrating its adaptability to increasing search space complexity. In addition to achieving state-of-the-art performance on benchmark datasets (GTEA Gaze, GTEA Gaze+, EPIC-Kitchens, and Charades-Ego), we establish a clear taxonomy for open-world recognition, delineating the challenges and methodological advancements necessary for egocentric activity understanding. Our results highlight the importance of structured search strategies, paving the way for scalable and efficient open-world activity recognition.
Code (in supplementary) will be shared publicly after review. 
\end{abstract}